\documentclass[sigconf]{acmart}
\renewcommand\footnotetextcopyrightpermission[1]{} 

\usepackage{fancyhdr}

\usepackage{colortbl}  
\usepackage{xcolor}

\usepackage[misc]{ifsym}
\usepackage[marginal]{footmisc}

\AtBeginDocument{%
  \providecommand\BibTeX{{%
    \normalfont B\kern-0.5em{\scshape i\kern-0.25em b}\kern-0.8em\TeX}}}

\acmConference{MM'23}{}{October 29 -- November 3, 2023, Ottawa, Canada.}

\acmSubmissionID{1099}

\usepackage{multirow}
\usepackage{enumitem}

\begin{document}

\title{Feature-Suppressed Contrast for Self-Supervised Food Pre-training}

%
%
\author{Xinda Liu$^*$}
\email{liuxinda@buaa.edu.cn}
\affiliation{%
 \institution{School of Computer Science and Engineering, Beihang University}
 \streetaddress{No.37, Xueyuan Road, Haidian District, Beijing, P.R.China.}
 \city{Beijing}
 \country{China}
 \postcode{100191}
}
\author{Yaohui Zhu$^*$}
\email{yaohui.zhu@bnu.edu.cn}
\affiliation{%
 \institution{School of Artificial Intelligence, Beijing Normal University}
 \streetaddress{No.19, Xinjiekouwai Street, Haidian District, Beijing, P.R.China.}
 \city{Beijing}
 \country{China}
 \postcode{100191}
}

\author{Linhu Liu}
\email{liulh7@lenovo.com}
\affiliation{%
 \institution{AI Lab, Lenovo Research}
 \streetaddress{No.10. Courtyard Xibeiwang East Road Haidian District, Beijing, China}
 \city{Beijing}
 \country{China}
 \postcode{100094}
}

\author{Jiang Tian}
\email{tianjiang1@lenovo.com}
\affiliation{%
 \institution{AI Lab, Lenovo Research}
 \streetaddress{No.10. Courtyard Xibeiwang East Road Haidian District, Beijing, China}
 \city{Beijing}
 \country{China}
 \postcode{100094}
}

\author{Lili Wang \textsuperscript{\Letter}}
\email{wanglily@buaa.edu.cn}
\affiliation{%
 \institution{School of Computer Science and Engineering, Beihang University}
 \streetaddress{No.37, Xueyuan Road, Haidian District, Beijing, P.R.China.}
 \city{Beijing}
 \country{China}
 \postcode{100191}
}

\thanks{$^*$ Both authors contributed equally to this research. Lili Wang is the corresponding author.}


\label{abstract}
\begin{abstract}

Most previous approaches for analyzing food images have relied on extensively annotated datasets, resulting in significant human labeling expenses due to the varied and intricate nature of such images. Inspired by the effectiveness of contrastive self-supervised methods in utilizing unlabelled data, we explore leveraging these techniques on unlabelled food images. In contrastive self-supervised methods, two views are randomly generated from an image by data augmentations. However, regarding food images, the two views tend to contain similar informative contents, causing large mutual information, which impedes the efficacy of contrastive self-supervised learning. To address this problem, we propose Feature Suppressed Contrast (FeaSC) to reduce mutual information between views. As the similar contents of the two views are salient or highly responsive in the feature map, the proposed FeaSC uses a response-aware scheme to localize salient features in an unsupervised manner. By suppressing some salient features in one view while leaving another contrast view unchanged, the mutual information between the two views is reduced, thereby enhancing the effectiveness of contrast learning for self-supervised food pre-training. As a plug-and-play module, the proposed method consistently improves BYOL and SimSiam by 1.70\% $\sim$  6.69\% classification accuracy on four publicly available food recognition datasets. Superior results have also been achieved on downstream segmentation tasks, demonstrating the effectiveness of the proposed method.

\end{abstract}




\maketitle

\section{Introduction} \label{sec.intro}

Food image analysis involves the utilization of visual data to determine food attributes, including quality, quantity, composition, and nutrient content. It has the potential to contribute to various applications, such as dietary assessment \cite{jiang2020deepfood, wang2022review}, food inspection \cite{chen2021review,brosnan2004improving}, food recognition \cite{min2023large,zhu2023learn}, and food recommendation \cite{min2019food}. Furthermore, it raises awareness regarding eating habits, reduces food waste, promotes food diversity, and ensures food safety.
 The majority of food image analysis methods \cite{min2023large,zhu2023learn} rely on large-scale annotated datasets.
However, it is challenging to annotate such large-scale datasets due to the diverse and complex food images across different regions, influenced by natural conditions and cultural disparities. 
In certain cultures, foods with the same name may be prepared differently, causing difficulties in the annotation process.
Therefore, utilizing unlabelled food images for analysis is significant and valuable in the multimedia community.

Self-supervised learning methods, specifically contrastive self-supervised methods \cite{caron2020unsupervised, he2020momentum, grill2020bootstrap, chen2021exploring},  have provided a promising solution for utilizing unlabelled data. 
In the realm of food image analysis, self-supervised learning presents a novel perspective on solving its challenges. 
The concept behind contrastive self-supervised methods is to compare different views of a single image in order to derive invariant feature representations.
The methodology's fundamental principle is to enhance feature consistency between different views of the same sample whilst retaining discrepancies between different samples.
Pre-training on ImageNet dataset enables these methods to achieve comparable performance to supervised learning techniques. Currently, self-supervised learning mainly focuses on ImageNet images, and there has been a conspicuous scarcity of analysis into self-supervised learning concerning food images. Therefore, this paper aims to explore self-supervised learning on food images.


\begin{figure}[t]
  \centering
  \includegraphics[width=0.99\linewidth]{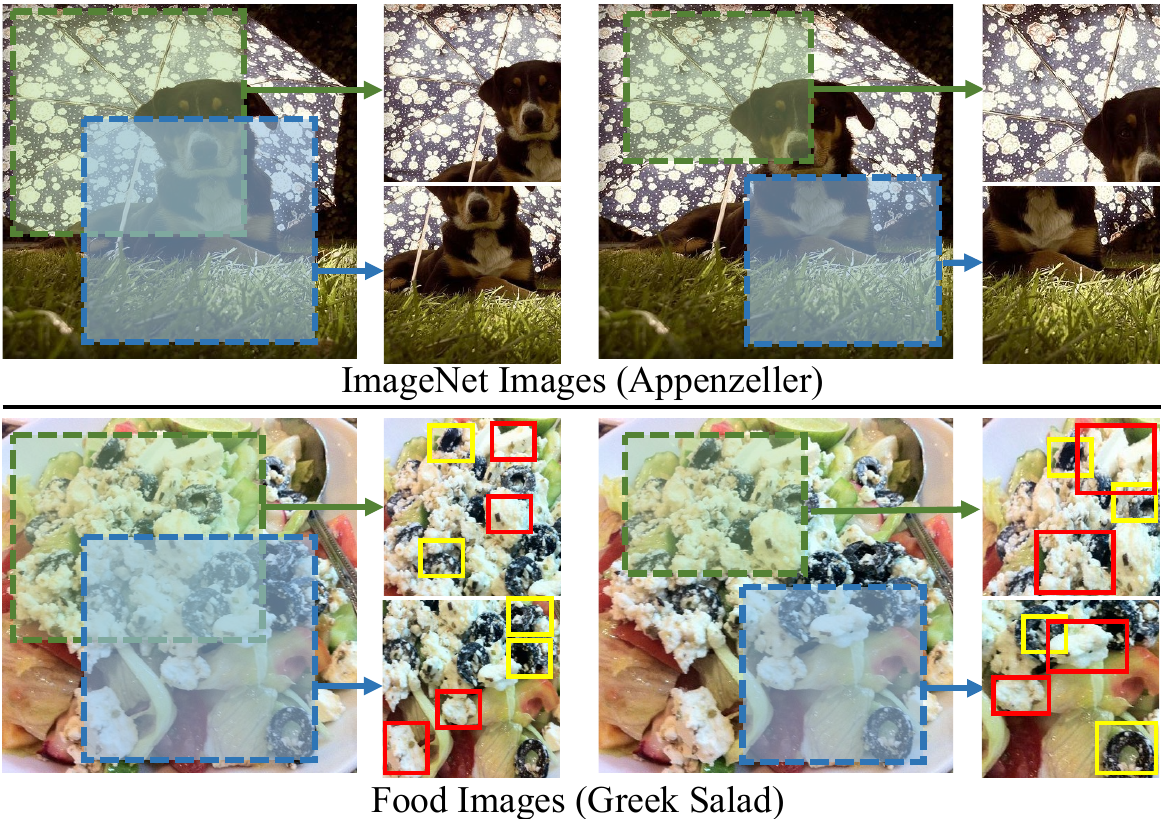}
   \caption{Some examples of  views sampled from Appenzeller images and Greek Salad images. The views of Appenzeller easily capture different object parts, while the views of Greek Salad tend to contain similar informative contents.}
   \label{fig:intro_ex1}
   \vspace{-1.5em}
\end{figure}

Compared with ImageNet images, food images include a variety of similar ingredients stacked together. As shown in Figure \ref{fig:intro_ex1}, there are different semantic parts (e.g., head and body) in ’Appenzeller’, while lots of sliced material (e.g., cucumbers and cheese), fruit and vegetable are stacked together in Greek Salad’. 
In contrastive self-supervised methods, each image is randomly sampled to form two different views by data transformation operations. 
On ImageNet images, the views easily capture different object parts, while the views of food images tend to contain similar informative  contents.
As stated in \citep{tian2020makes,lioptimal}, good views are made by reducing the mutual information between
them while keeping task-relevant information intact. 
We think two views with similar informative contents keep task-relevant information, however, their similarity result in large mutual information between them, which prevents effective contrastive self-supervised learning. 
Based on the above analyse, we argue that it is feasible to reduce some similar informative contents between two views for contrastive self-supervised food learning.

\begin{figure}[t]
  \centering
  \includegraphics[width=0.98\linewidth]{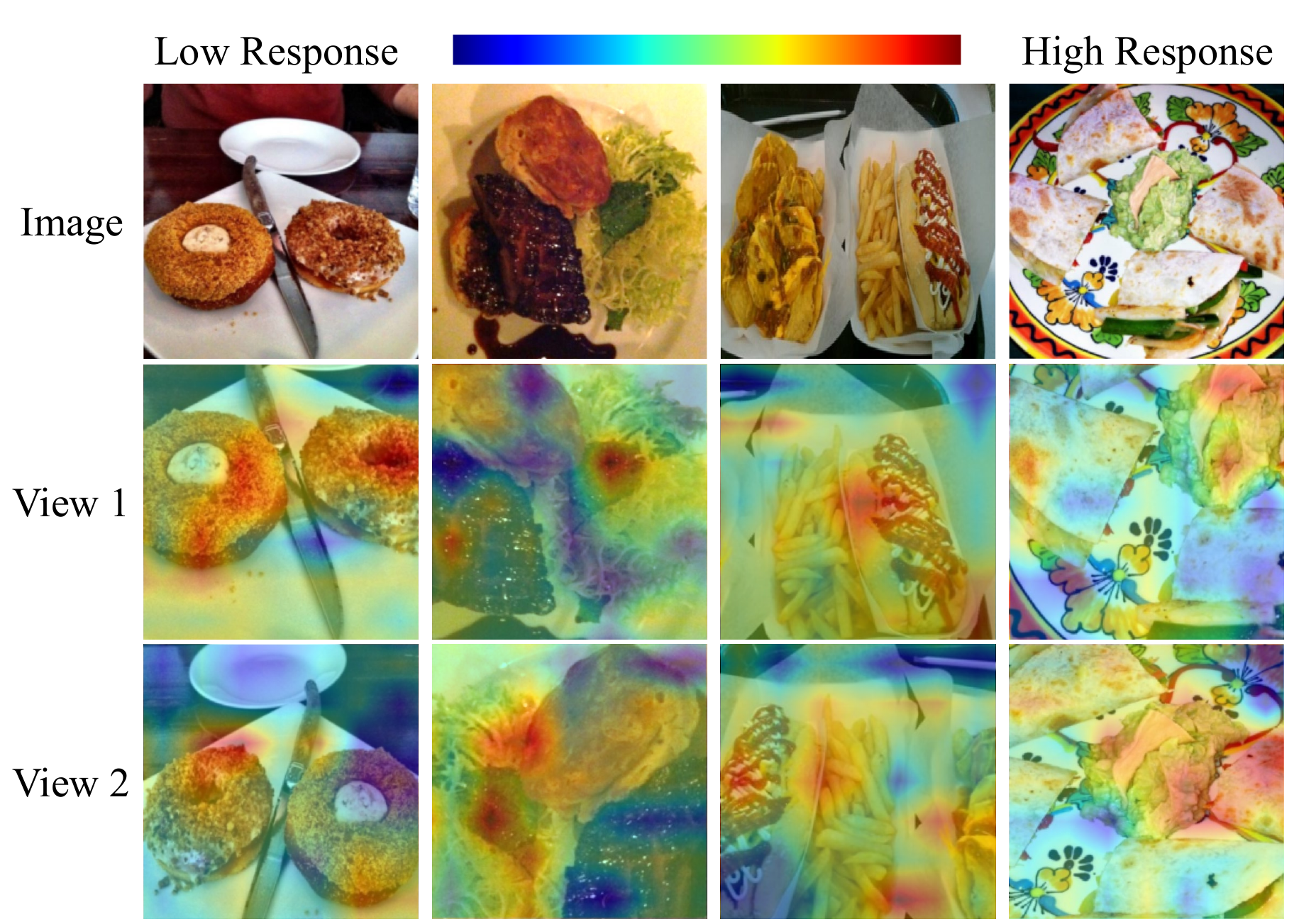}
  \caption{Some examples of images and corresponding views. The similar contents in the two views of the same image are both highly responsive.}
  \label{fig:intro_ex2}
\end{figure}


To this end, based on contrastive self-supervised learning, we propose Feature-Suppressed Contrast (FeaSC) to boost self-supervised food pre-training via excluding comparisons of similar informative contents. 
We observe that the similar contents of the two views are salient or highly responsive in feature map, as shown in Figure \ref{fig:intro_ex2}. Here, we use an unsupervised approach, similar to \citep{wei2017selective,choe2019attention,shen2022semicon}, instead of the category-dependent supervised approach of CMA \citep{zhou2016learning, selvaraju2017grad, cheng2023class}.
Meanwhile, the view with some salient feature suppressed  still maintains task-relevant information \citep{zhu2023learn}, since the view of food images contains many informative contents stacked together.
Therefore, we introduce a response-aware scheme to localize salient features in an unsupervised manner.
The proposed FeaSC suppresses some salient features in one view,  while leaving another contrast view unchanged. 
In this way, the mutual information between the two views is reduced, which enhances the effectiveness of contrast learning for self-supervised food pre-training.
As a plug-and-play method, it can be easily applied to different contrastive self-supervised frameworks. 
The proposed method consistently improves BYOL \citep{grill2020bootstrap}, SimSiam \citep{chen2021exploring} by 1.70\% $\sim$  6.69\% classification accuracy on ETHZ Food-101 \citep{Bossard-Food101-ECCV2014}, Vireo Food-172 \citep{Chen-DIRCRR-MM2016}, ISIA Food-200 \citep{min2019ingredient}, and ISIA Food-500 \citep{min2020isia}, under linear evaluations. Notably, when  10\% training data is employed under linear evaluations, performance improvements of our method are 4.37\% $\sim$ 20.96\%  on the four datasets. 
Moreover, superior results are also been achieved on downstream food segmentation tasks.

The main contributions of this paper are summarized as:
\begin{itemize}
\item We propose a feature-suppressed contrast method to boost self-supervised food pre-training via excluding comparisons of similar informative contents.
\item We further propose a response-aware localization scheme to improve the efficiency of feature suppression.
\item Extensive experimental evaluations on several public food downstream tasks demonstrate the effectiveness of the proposed method.
\end{itemize}


\section{Related Works} \label{sec.rel}

\subsection{Contrastive Self-Supervised Learning}

As a form of unsupervised learning, contrastive self-supervised methods \cite{caron2021emerging,chen2020simple, he2020momentum,grill2020bootstrap, yan2022repeatable, sun2023enhancing} have demonstrated superior abilities in learning generalizable representations. The core idea of contrastive self-supervised methods is to maximize feature consistency under different views from the same instance, while pushing features of different instances apart. 
According to whether negative instances are used, contrastive self-supervied methods are divided into two categories.

One of them is to compare different views sampled from both positive instances and negative instances with the InfoNCE loss \citep{oord2018representation}. In these methods, negative instances play a critical role and are carefully designed. 
Chen et al. \cite{chen2020simple} point out that contrastive learning benefits from a large number of instances in a batch comparison and composition of data augmentations. He et al. \cite{he2020momentum}
leverage a dynamic dictionary with a queue and a moving-averaged encoder to provide consistent representations of negative instances on-the-fly. These methods only update the representation of samples from the current batch, possibly discarding the useful information from the past batches. Alternatively, He et al. \cite{hu2021adco} propose to directly learn a set of negative adversaries playing against the self-trained representation. In addition,  some methods combine contrastive learning with clustering \citep{caron2020unsupervised,liprototypical} or adversarial training \citep{kim2020adversarial,fan2021does}.

Another category is to compare different views sampled from the same instance. One crucial issue in this kind of method is model collapse, where all data is mapped to the same representation. Grill et al. \citep{grill2020bootstrap} first rely only on positive pairs for contrast learning via
a momentum encoder and a stop-gradient operation.
Chen et al. \citep{chen2021exploring} conclude that the stop-gradient operation is critical to preventing mode collapse. In the above two works, the asymmetrical network architecture is implemented with a unique predictor, and the parameter updates involve a stop-gradient operation or with a momentum encoder in an asymmetrical manner.
Different from them, some works employ symmetric architectures and parameter updates via redundancy reduction \citep{zbontar2021barlow}, feature decorrelation \citep{hua2021feature}, and variance-invariance covariance regularization \citep{bardes2022variance}.
In this paper, based on these methods, we propose a feature-suppressed contrast method for self-supervised food pre-training.

\begin{figure*}[ht]
  \centering
    \includegraphics[width=\linewidth]{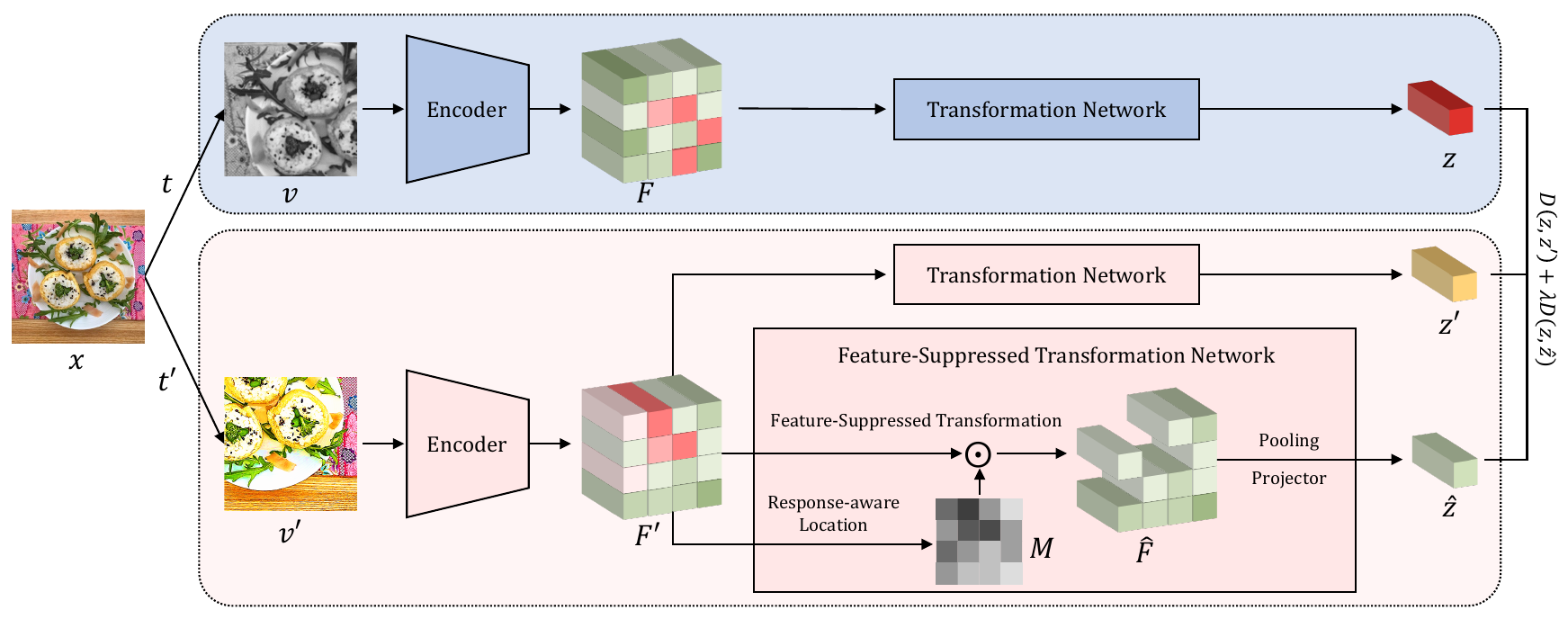}

   \caption{The pipeline of the proposed FeaSC. Two different views are randomly sampled from an image, and then they are respectively inputted into an encoder to obtain their feature map via two branches.  The feature map in bottom branch inputs a transformation network and a feature-suppressed transformation network to obtain two feature representations, which contrast a feature representation generated from the top branch.}
   \label{fig:framework}
\end{figure*}

\subsection{Data Transformation}
The commonly used data transformation (augmentation) involves spatial/geometric transformation such as cropping and resizing, and appearance transformation such as color distortion and Gaussian blur. In contrast self-supervised learning, Chen et al. \cite{chen2020simple} point out that it is crucial to composite multiple data augmentation operations. Further, Tian et al. \citep{tian2020makes} analyze the influence of different views of data transformation, and argue that a better pre-training model can be obtained by reducing the mutual information between views while keeping task-relevant information intact. Peng et al. \citep{peng2022crafting} propose ContrastiveCrop to generate better views. And some works \citep{shi2022adversarial,assran2022masked} propose to mask views on image levels under the contrast learning framework. 
In addition, some self-supervised methods of masked image modeling \cite{bao2021beit,zhou2022ibot,wei2022masked,he2022masked,xie2022simmim} randomly mask out some input image tokens and then recover the masked content by conditioning on the visible context. The data transformation of the above works is on  image levels. Different from these works, we introduce data transformation on the feature map, which is effective for self-supervised food pre-training.

\subsection{Food Pre-training}
The majority of food-related works use a model pre-trained on ImageNet dataset to initialize their models, such as food recognition \citep{min2020isia,zhu2023learn} and food category-ingredient prediction \citep{wang2022ingredient}.
This is mainly because the fact that there are no food pre-training models available.
Recently, Min et al. \citep{min2023large} propose a large-scale dataset of food recognition and demonstrate that a pre-trained model on this dataset brings more significant benefits for food-related downstream tasks than the model pre-trained on ImageNet dataset. This gives us a message that it is meaningful to study the food pre-trained model on a large food dataset. Therefore, in this paper, we study the food pre-trained model on this large dataset in a self-supervised manner. We hope our explorations are helpful to provide food-related research with a better pre-trained model.



\section{The Proposed Method} \label{sec.method}

The proposed FeaSC suppresses informative features in one view to avoid comparisons of similar contents between two different views.
As shown in Figure \ref{fig:framework}, the proposed method consists of two branches that process two different views randomly sampled from the input image. The top branch is commonly used, while the bottom branch is transformed by a feature-suppressed network that suppresses informative features. In this section, we first revisit contrastive self-supervised learning and then introduce how to suppress informative features and calculate the involved contrastive loss. Finally, we discuss the favorable properties of our method for better understanding.


\subsection{Revisiting Contrastive Self-Supervised Learning}
Contrastive self-supervised learning aims to learn generalized representations that are invariant to data augmentations by attracting positive pairs and repelling negative pairs in a latent space. Typically,  contrastive self-supervised methods are based on a siamese framework. Next, we will review contrastive self-supervised methods from the following three aspects: feature extraction, feature 
transformation, feature contrast. 


\textbf{Feature Extraction.} Given a set of images  $\mathcal{D}$, an image  $x$ sampled  from  $\mathcal{D}$ to produce two views $v \triangleq t(x)$  and  $v^{\prime} \triangleq t^{\prime}(x)$ by using random data transformations (i.e., compositions of  image augmentations), where $t()$ and $t^{\prime}()$ are two different data transformations with their specific image augmentations. Then the two views $v$ and $v^{\prime}$ are inputted an encoder network $f(;)$ to obtain their feature map $F$ and  $F^{\prime}$, respectively, which are formalized as follows:
\begin{equation} \label{equ.encoder}
                \left\{\begin{matrix}
                       F = f(v;\theta) \\
                       F^{\prime} = f(v^{\prime};\theta^{\prime})   
                       \end{matrix}\right.
\end{equation}
where $\theta$ and $\theta^{\prime}$ are parameters of the encoder network. The settings of two parameters differ in different contrastive self-supervised methods, for example, they are the same in SimSiam \citep{chen2021exploring}, Barlow twins \citep{zbontar2021barlow} and VICreg \citep{bardes2022variance}, while they are different in BYOL \citep{grill2020bootstrap} and DINO \citep{caron2021emerging}.

\textbf{Feature Transformation.} Subsequently,  the two feature map $F$ and  $F^{\prime}$ are inputted into a transformation network to get their feature representations $z$ and  $z^{\prime}$, respectively. This process is formalized as follows:
\begin{equation} \label{equ.transformation}
    \left\{\begin{matrix}
       z=g(F;\xi) \\
       z^{\prime} =g^{\prime}(F^{\prime};\xi^{\prime}) 
       \end{matrix}\right.
\end{equation}
where $\xi$ and $\xi^{\prime}$ are parameters of the transformation network $g(;)$ and $g^{\prime}(;)$, respectively. The architectures of the two transformation network differ in different contrastive self-supervised methods. In DINO, the architectures and parameters of the two transformation network are the same, which contain a pooling operation, a MLP projection layer and a projector. 
In Barlow twins and VICreg, the architectures and parameters of the two transformation network are also the same, but they contain a pooling operation and a projector. 
In BYOL and SimSiam, one transformation network contains a pooling operation and  a MLP projection layer, another one contains a pooling operation, a MLP projection layer, and a projector.

\textbf{Feature Contrast.} The final optimization goal is to minimize distance between feature representations $z$ and  $z^{\prime}$, namely:
\begin{equation} \label{equ.original_loss}
\mathcal{L} = D(z,z^{\prime})
\end{equation}
The optimization loss is slightly different in different contrastive self-supervised methods. 
In DINO, the optimization goal is a cross-entropy loss.
In BYOL, the goal is mean-squared euclidean distance between  normalized $z$ and  $z^{\prime}$. In SimSiam, the goal is negative cosine similarity between $z$ and  $z^{\prime}$.

\subsection{Feature-Suppressed Contrast}

Feature-suppressed contrast includes three key components response-aware location, feature-suppressed transformation, and the final contrast module. The response-aware localization module identifies the salient regions of the feature map. According to the localization of salient regions, the feature-suppressed transformation module modifies the feature map to suppressed representations, which are  used in final contrast module. Next, we introduce the three components.

\textbf{Response-Aware Location.} The salient contents can be located with high class responses of feature map.  Given a feature map $F \in \mathbb{R}^{C \times W \times H}$, where $C$, $H$, and $W$ denote the number of channels, height, and width, respectively, the response values are summed over the channel dimension with Eq. \ref{equ.sum},
\begin{equation} \label{equ.sum}
    M_{i,j} =  \sum_{k=1}^{C}F_{i,j,k}
\end{equation}
where $M_{i,j}$ represents the $i^{th}$ row and the $j^{th}$ column in $M \in  \mathbb{R}^{ W \times H}$.
The areas marked with high value in the response map $M$ indicate salient regions.
Therefore, these regions can be located by utilizing such high values.
 A certain percentile $\eta \in \left [ 0,1 \right ]$ in $M$ is used as a threshold. Supposing the value of the percentile is $\omega$, the location of regions to be suppressed is calculated as:
\begin{equation} \label{equ.loc}
    Loc_{i,j} = \left\{\begin{matrix}
                 1,& M_{i,j} \geq \omega \\ 
                 0,& M_{i,j} \leq \omega
                \end{matrix}\right.
\end{equation}
 where $M_{i,j} \in M$.
To maintain the stability of the training, we use a  ramp-up function to determine the percentage $\eta$ of feature suppression. The $\eta$ starts from a small value to a fixed value $\alpha$, and its formulation is as follows:
\begin{equation} \label{equ.alpha}
    \eta=\left\{\begin{array}{ll}
\alpha* \exp \left(-5\left(1-\frac{e}{\beta}\right)^{2}\right), & e<\beta \\
\alpha, & e \geq \beta
\end{array}\right.
\end{equation}
where $e$ denotes the current epoch during training phase, $\alpha$ is a scalar, and $\beta$ is an integer.

At the begin of training, the capacity of self-supervised model is not strong.  A small value of $\eta$ can reduce tough contrast between some views to improve effective learning of a weak self-supervised model.
After the capacity of self-supervised model become enough strong, a big value of $\eta$ can encourage the model to mine other informative features.

\textbf{Feature-Suppressed Transformation.} 
The location of the salient regions $Loc_{i,j}$ is used to suppress the corresponding features.
The formulation of this process is defined as follows:
\begin{equation} \label{equ.FST}
    \hat{F}_{i,j,k} = (1-Loc_{i,j}) \odot F_{i,j,k}
\end{equation}
where $\odot$ means an element-wise multiplication, and $\hat{F}$ is the suppressed feature map. Then suppressed feature map $\hat{F}$ is pooled into a feature vector. Subsequently, the feature vector inputs a projector to get feature representations $\hat{z}$.

The feature representations $\hat{z}$ lack some salient information since the highly responsive features are suppressed in the feature map. The feature-suppressed transformation is carried out in one view, while another view does not. 
Therefore, the comparisons of salient similar contents between the two views are avoided, achieving a reduction in mutual information between the two views. Moreover, comparing different contents between the two views is more likely to encourage the model to mine other distinctive and informative features in the suppressed views.

\textbf{Final Contrast.}
The final contrast loss consists of two terms, which is defined as follows:
\begin{equation} \label{equ.final_loss}
    \mathcal{L} = D(z, z^{\prime}) + \lambda D(z, \hat{z})
\end{equation}
where $D(z, z^{\prime})$ is an original contrast loss, $D(z, \hat{z})$ is a contrast loss on suppressed features, and $\lambda$ is a hyper-parameter to balance the two terms. 
By preserving the original contrast loss, more precise responsive localization can be retained, facilitating effective contrast learning for food pre-training model.

\subsection{Discussions}



\begin{figure}[h]
\setlength{\abovecaptionskip}{0cm}
  \centering
   \includegraphics[width=0.99\linewidth]{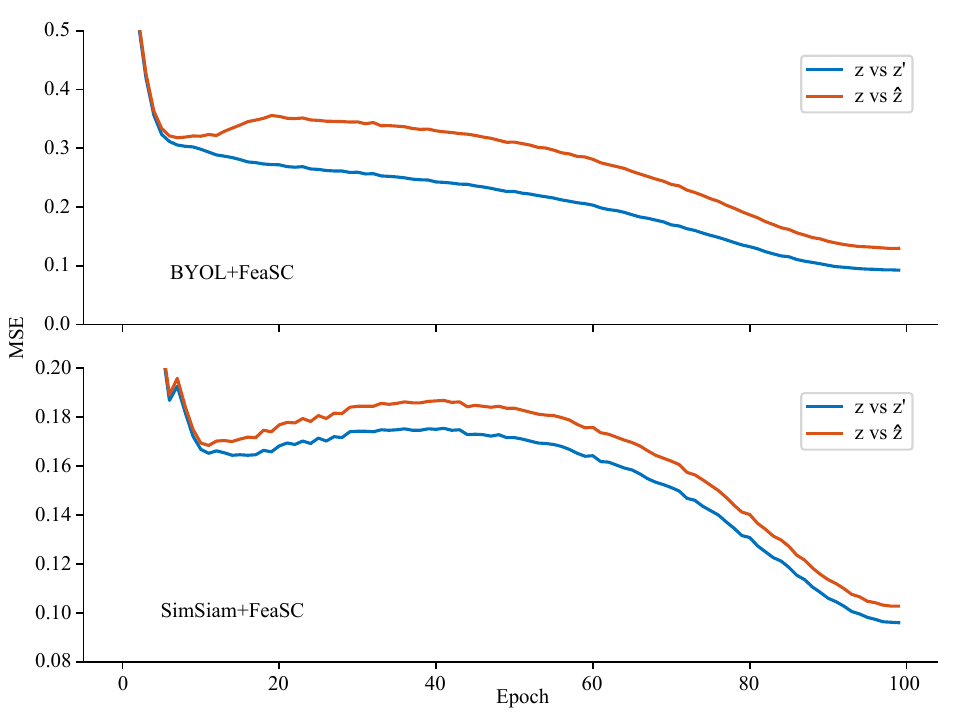}

   \caption{Visualization of the MSE obtained by comparison of views with and without suppression.
   }
   \label{fig:loss}
   \vspace{-1.5 em}
\end{figure}

\textbf{Relation with reducing mutual information.}
Mutual information between $z$ and $z^{\prime}$ can be calculated using the following formula:
\begin{equation} \label{equ.mse1}
\begin{aligned}
    \mathcal{I}(z,z^{\prime}) &= \mathcal{H}(z) - \mathcal{H}(z|z^{\prime}) = \mathcal{H}(z) + \mathbb{E}_{p}\log p(z|z^{\prime}) \\
    &\geq \mathcal{H}(z) + \mathbb{E}_{p}\log q(z|z^{\prime}) \\
    &= \mathcal{H}(z) - 1/2 \log \mathbb{E}_{p}[(z-z^{\prime})^2] - C
\end{aligned}
\end{equation}
$\mathcal{H}(z)$ is the entropy of $z$, and $\mathcal{H}(z|z^{\prime})$ is the conditional entropy of $z$ given $z^{\prime}$.
The inequality in the second line of the formula is derived from Gibbs' inequality. Letting $q(z|z^{\prime}) \sim \mathcal{N}(\mu_{z|z^{\prime}}, \mathbb{E}_{p}[(z-z^{\prime})^2])$, it gives the equation in the third row, where $C= 1/2\log(2\pi)+1/2$.
Similarly, the following inequality is obtained:
\begin{equation} \label{equ.mse2}
    \mathcal{I}(z, \hat{z}) \geq  \mathcal{H}(z) - 1/2 \log \mathbb{E}_{p}[(z-\hat{z})^2] - C
\end{equation}
From Eq. \ref{equ.mse1} and Eq. \ref{equ.mse2}, the lower bound of their mutual information (i.e., $\mathcal{I}(z,z^{\prime})$ and $\mathcal{I}(z, \hat{z})$) can be estimated by their mean-square error (MSE) (i.e., $(z-z^{\prime})^2$ and $(z-\hat{z})^2$). The larger the MSE, the smaller lower bound of the mutual information. The comparisons of the two MSE on both BYOL and SimSiam methods are shown in Figure \ref{fig:loss}. The MSE between $z$ and $\hat{z}$ is larger than it between between $z$ and $z^{\prime}$.
Therefore, we can obtain that $\mathcal{I}(z, \hat{z})$ has a lower bound than $\mathcal{I}(z,z^{\prime})$.
This is to say, suppressing feature increases the MSE while raising a lower bound of the mutual information. This is also consistent with the intuition that mutual information decreases when the similarity of features in different views decreases.

\section{Experiments} \label{sec.exp}

To evaluate the effectiveness of our proposed method, we follow standard practice on a series of downstream tasks, including food recognition and food segmentation. 

\subsection{Self-Supervised Settings}

We plug the proposed FeaSC into two contrastive self-supervised learning frameworks: SimSiam \citep{chen2020simple} and BYOL \citep{grill2020bootstrap}, forming the corresponding methods SimSiam+FeaSc and BYOL+FeaSC. 
For a fair comparison,  all models using ResNet-50 as the backbone are trained for 100 epochs with a batch size of 512 during the self-supervised learning phase. This process takes under three days to converge using four NVIDIA V100 GPUs.
The SGD optimizer is utilized with momentum set to 0.5 and weight decay set to 1e-4. The learning rate follows a cosine decay schedule from 0.5 to 0, with 20 warm-up epochs. The augmentation configurations are identical to those used in SimSiam. $\alpha$ and $\beta$ in Eq. \ref{equ.alpha} is set to 0.2 and 20, respectively.
The networks are pre-trained on Food2K \citep{min2023large} or ImageNet-1K.

\begin{table*}[h]
\centering
\begin{tabular}{ccccccccccc}
\toprule
                         &                                                                           &                              & \multicolumn{4}{c}{ETHZ Food-101}                                                                                         & \multicolumn{4}{c}{Vireo Food-172} \\ \cmidrule(lr){4-7} \cmidrule(l){8-11} 
\multirow{-2}{*}{Method} & \multirow{-2}{*}{\begin{tabular}[c]{@{}c@{}}Pretrain\\ Data\end{tabular}} & \multirow{-2}{*}{Evaluation} & 10\%                        & 20\%                          & 50\%                          & 100\%                            & 10\%     & 20\%    & 50\%    & 100\%      \\ \toprule 
Supervised*               & ImageNet-1K                                                               & Linear                       & 60.37                        & 74.02                        & 77.74                        & 79.90                         & 68.52   & 73.57  & 78.44  & 81.00     \\
Supervised*              & Food2K                                                                    & Linear                       & 63.83                        & 67.80                         & 71.39                        & 73.49                        & 78.04   & 79.58  & 81.75  & 83.17  \\ 
BYOL*                  & ImageNet-1K                                                               & Linear                       & 63.61                        & 68.58                        & 72.42                        & 74.83                        & 68.54   & 73.61  & 78.37  & 80.95  \\
BYOL                     & Food2K                                                                    & Linear                       & 68.31                        & 72.64                        & 76.77                        & 79.66                        & 72.57   & 77.76  & 82.24  & 84.59  \\
\rowcolor{green!20}BYOL+FeaSC                 & Food2K                                                                    & Linear                       & 72.69                        & 75.92                        & 79.39                        & 81.36                         & 81.83   & 83.99  & 86.33  & 87.86  \\ 
SimSiam*                & ImageNet-1K                                                               & Linear                       & 29.60                         & 31.20                         & 34.11                        & 39.26                        & 12.49   & 14.84  & 21.08  & 26.96  \\
SimSiam                  & Food2K                                                                    & Linear                       & 54.01                        & 59.66                        & 66.08                        & 70.10                         & 50.31   & 59.46  & 71.37  & 76.89  \\
\rowcolor{green!20}SimSiam+FeaSC              & Food2K                                                                    & Linear                       & 61.22                        & 65.66                        & 70.26                        & 73.52                        & 71.27   & 75.44  & 79.56  & 81.58  \\ 
\toprule
Supervised*              & ImageNet-1K                                                               & Fine-tuning                  & 70.95                        & 76.86                        & 82.51                        & 86.16                        & 71.75   & 77.58  & 83.79  & 87.28  \\
Supervised*              & Food2K                                                                    & Fine-tuning                  & 74.67                        & 79.03                        & 83.80                         & 87.09                        & 81.56   & 83.91  & 86.97  & 88.91  \\ 
BYOL*                   & ImageNet-1K                                                               & Fine-tuning                  & 67.76                        & 74.52                        & 80.85                        & 84.64                        & 68.04   & 74.90   & 81.99  & 85.90   \\
BYOL                     & Food2K                                                                    & Fine-tuning                  & 77.03                        & 81.68                        & 85.85                        & 88.25                        & 81.69   & 84.58  & 87.93  & 89.68  \\
\rowcolor{green!20}BYOL+FeaSC                 & Food2K                                                                    & Fine-tuning                  & {77.83} & {81.83} & {86.21} & {88.39} & 84.20   & 86.59  & 88.88  & 90.54  \\ 
SimSiam*                 & ImageNet-1K                                                               & Fine-tuning                  & 63.87                        & 73.79                        & 82.19                        & 86.30                         & 63.98   & 74.85  & 82.91  & 86.61  \\
SimSiam                  & Food2K                                                                    & Fine-tuning                  & 73.99                        & 80.07                        & 85.30                         & 87.79                        & 79.14   & 83.30   & 86.99  & 89.22  \\
\rowcolor{green!20}SimSiam+FeaSC              & Food2K                                                                    & Fine-tuning                  & 75.37                        & {80.84} & {85.55} & {88.20} & 81.22   & 83.84   & 87.05  & 89.36  \\ \bottomrule

\end{tabular}
\caption{Top-1 accuracy (\%) of the different methods with four proportions on ETHZ Food-101 and Vireo Food-172. * indicates using officially provided pre-training parameters while other methods are our own implementations.}
\label{tab:recog1}
\vspace{-1.5em}
\end{table*}

\begin{table*}[h]
\begin{tabular}{ccccccccccc}
\toprule
                         &                                                                           &                              & \multicolumn{4}{c}{ISIA Food-200}                                                                                         & \multicolumn{4}{c}{ISIA Food-500} \\ \cmidrule(lr){4-7} \cmidrule(l){8-11} 
\multirow{-2}{*}{Method} & \multirow{-2}{*}{\begin{tabular}[c]{@{}c@{}}Pretrain\\ Data\end{tabular}} & \multirow{-2}{*}{Evaluation} & 10\%                          & 20\%                          & 50\%                          & 100\%                            & 10\%                          & 20\%                          & 50\%                          & 100\%                            \\  \toprule 
Supervised*               & ImageNet-1K                                                               & Linear                       & 47.78                        & 51.58                        & 56.09                        & 58.63                        & 41.38                        & 46.01                        & 50.80                         & 53.71                        \\
Supervised*               & Food2K                                                                    & Linear                       & 46.07                        & 49.37                        & 52.77                        & 54.86                        & 38.72                        & 42.33                        & 46.17                        & 48.59                        \\ 
BYOL*               & ImageNet-1K                                                               & Linear                       & 43.05                        & 47.15                        & 51.51                        & 53.89                        & 36.63                        & 40.91                        & 45.61                        & 48.39                        \\
BYOL                     & Food2K                                                                    & Linear                       & 44.08                        & 50.70                         & 56.08                        & 59.06                        & 33.18                        & 41.63                        & 48.76                        & 52.55                        \\
\rowcolor{green!20}BYOL+FeaSC                 & Food2K                                                                    & Linear                       & 53.21                        & 56.25                        & 59.93                        & 64.04                        & 45.36                        & 49.22                        & 53.35                        & 55.94                        \\ 
SimSiam*              & ImageNet-1K                                                               & Linear                       & 5.66                         & 9.30                          & 14.57                        & 18.18                        & 2.71                         & 4.46                         & 8.33                         & 11.26                        \\
SimSiam                  & Food2K                                                                    & Linear                       & 26.13                        & 33.08                        & 43.14                        & 49.02                        & 16.94                        & 22.51                        & 33.01                        & 40.32                        \\
\rowcolor{green!20}SimSiam+FeaSC              & Food2K                                                                    & Linear                       & 40.93                        & 46.06                        & 51.11                        & 53.99                        & 30.61                        & 37.03                        & 43.31                        & 47.01                        \\  \toprule 
Supervised*             & ImageNet-1K                                                               & Fine-tuning                  & 47.92                        & 53.60                         & 59.53                        & 63.56                        & 39.57                        & 46.3                         & 53.15                        & 57.94                        \\
Supervised*             & Food2K                                                                    & Fine-tuning                  & 52.64                        & 56.38                        & 61.50                         & 64.59                        & 45.26                        & 49.43                        & 54.85                        & 58.83                        \\ 
BYOL*               & ImageNet-1K                                                               & Fine-tuning                  & 46.45                        & 51.97                        & 58.61                        & 62.48                        & 39.82                        & 45.88                        & 52.67                        & 56.89                        \\
BYOL                     & Food2K                                                                    & Fine-tuning                  & 55.62                        & 59.10                         & 63.53                        & 66.17                        & 43.09                        & 51.97                        & 56.64                        & 60.07                        \\
\rowcolor{green!20}BYOL+FeaSC                 & Food2K                                                                    & Fine-tuning                  & {56.57} & {59.86} & {63.95} & {66.72}  & 48.57                        & 52.71                       & 57.43                        & 60.73                         \\ 
SimSiam*             & ImageNet-1K                                                               & Fine-tuning                  & 43.70                         & 51.57                        & 58.83                        & 64.69                        & 39.07                        & 46.75                        & 54.75                        & 59.54                        \\
SimSiam                  & Food2K                                                                    & Fine-tuning                  & 51.67                        & 56.55                        & 62.47                        & 65.63                        & 44.86                        & 50.13                        & 56.03                        & 60.04                        \\
\rowcolor{green!20}SimSiam+FeaSC              & Food2K                                                                    & Fine-tuning                  & 53.28                       & {57.86} & {62.82} & {66.29} & {46.02} & {50.90} & {56.98} & {60.62} \\ \bottomrule
\end{tabular}
\caption{Top-1 accuracy (\%) of the different methods with four proportions on  ISIA Food-200 and ISIA Food-500.}
\label{tab:recog2}
\vspace{-1.5em}
\end{table*}

\subsection{Food Recognition}

\textbf{Datasets.} Experiments are evaluated on four commonly used food recognition datasets, including ETHZ Food-101 \citep{Bossard-Food101-ECCV2014}, Vireo Food-172 \citep{Chen-DIRCRR-MM2016}, ISIA Food-200 \citep{min2019ingredient}, and ISIA Food-500 \citep{min2020isia}.
\textbf{ETHZ Food-101}  contains 101 food categories, Each of which has 1,000 images including 750 training images and 250 test images. 
\textbf{Vireo Food-172}  contains 172 categories with 110,241 images. 
\textbf{ISIA Food-200} contains 200 food categories, Each of which has at least 500 images including 750 training images and 250 test images. 
\textbf{ISIA Food-500}  consists of 399,726 images with 500 categories. The average number of images per category is about 800. 

\textbf{Evaluation Protocols.}  For the evaluation of the self-supervised model, there are two standard protocols.
The first involves training a linear classifier while keeping the pre-trained weights fixed, known as linear evaluation. The second consists in training the entire network parameters and initializing the backbone with the pre-trained weights, known as fine-tuning evaluation.  
Data augmentation techniques such as random resize crops, and horizontal flips are applied during training, while a central crop is used for inference. Additionally, partial training datasets are used under various settings (e.g., 10\%, 20\%, 50\%) to compare the pre-training model's generalization ability thoroughly.

\textbf{Experimental results.}
Table \ref{tab:recog1} and Table \ref{tab:recog2} show the experimental results on four public food recognition datasets. 
Two self-supervised methods pre-trained on Food2K significantly outperform their corresponding ones pre-trained on ImageNet-1K in both linear and fine-tuning evaluations.
For instance, when using the entire training data for linear evaluation, SimSiam pre-trained on Food2K outperforms it pre-trained on ImageNet-1K by an average of 35.17\% on four datasets: ETHZ Food-101 (30.84\%), Vireo Food-172 (49.93\%), ISIA Food-200 (30.84\%), and ISIA Food-500 (29.06\%).
These results are noteworthy because the Food2K dataset contains food images smaller than the 1.3 million generic images in ImageNet-1K. The findings suggest that there are significant differences between food images and generic images. Thereby, there is good potential for research in self-supervised pre-training of food images.
\begin{table*}[h]
\begin{tabular}{cccccccccc}
\toprule
\multirow{2}{*}{Method} & \multirow{2}{*}{\begin{tabular}[c]{@{}c@{}}Pretrain\\ Dataset\end{tabular}} & \multirow{2}{*}{\begin{tabular}[c]{@{}c@{}}Segmentation \\ Method\end{tabular}} & \multirow{2}{*}{Evaluation} & \multicolumn{3}{c}{     FoodSeg 103     } & \multicolumn{3}{c}{UEC-FoodPix Complete} \\ \cmidrule(lr){5-7} \cmidrule(l){8-10}
                        &                                                                          &                                                                                 &                             & aAcc     & mIoU     & mAcc     & aAcc        & mIoU        & mAcc       \\  \toprule 
Supervised*           & ImageNet-1K                                                              & DeeplabV3                                                                       & Head                      & 24.34    & 4.46     & 7.86     & 23.61       & 9.92        & 17.67      \\
Supervised*           & Food2K                                                                   & DeeplabV3                                                                       & Head                      & 24.09    & 4.05     & 7.43     & 21.98       & 8.61        & 16.34      \\ 
BYOL*           & ImageNet-1K                                                              & DeeplabV3                                                                       & Head                      & 12.68    & 0.88     & 2.13     & 9.21        & 1.44        & 4.00          \\
BYOL                    & Food2K                                                                   & DeeplabV3                                                                       & Head                      & 17.49    & 2.67     & 4.99     & 19.23       & 7.15        & 13.14      \\
\rowcolor{green!20}BYOL+FeaSC                & Food2K                                                                   & DeeplabV3                                                                       & Head                      & 21.08    & 3.72     & 6.71     & 24.00          & 10.46       & 19.07      \\ 
SimSiam*           & ImageNet-1K                                                              & DeeplabV3                                                                       & Head                      & 17.87    & 2.51     & 4.85     & 16.74       & 5.02        & 10.53      \\
SimSiam                 & Food2K                                                                   & DeeplabV3                                                                       & Head                      & 15.81    & 1.90      & 3.94     & 16.80        & 5.49        & 11.13      \\
\rowcolor{green!20}SimSiam+FeaSC             & Food2K                                                                   & DeeplabV3                                                                       & Head                      & 20.01    & 2.96     & 5.75     & 16.82       & 5.54        & 11.48      \\ \hline
Supervised*           & ImageNet-1K                                                              & FCN                                                                             & Head                      & 23.27    & 3.62     & 6.84     & 15.83       & 4.87        & 10.16      \\
Supervised*          & Food2K                                                                   & FCN                                                                             & Head                      & 21.89    & 3.16     & 6.36     & 14.77       & 4.62        & 9.68       \\ 
BYOL*            & ImageNet-1K                                                              & FCN                                                                             & Head                      & 12.51    & 0.76     & 2.04     & 8.36        & 0.85        & 2.38       \\
BYOL                    & Food2K                                                                   & FCN                                                                             & Head                      & 18.84    & 2.66     & 5.10      & 13.98       & 3.95        & 7.76       \\
\rowcolor{green!20}BYOL+FeaSC                & Food2K                                                                   & FCN                                                                             & Head                      & 21.37    & 3.44     & 6.35     & 16.56       & 5.63        & 11.02      \\ 
SimSiam*           & ImageNet-1K                                                              & FCN                                                                             & Head                      & 20.38    & 3.06     & 5.88     & 13.50        & 3.78        & 8.07       \\
SimSiam                 & Food2K                                                                   & FCN                                                                             & Head                      & 16.57    & 1.91     & 3.98     & 12.07       & 2.87        & 6.25       \\
\rowcolor{green!20}SimSiam+FeaSC             & Food2K                                                                   & FCN                                                                             & Head                      & 20.79     & 3.07     & 6.04     & 13.54       & 3.79        & 8.17       \\ \toprule 
Supervised*          & ImageNet-1K                                                              & DeeplabV3                                                                       & Fine-tuning                    & 62.96    & 35.02    & 47.27    & 79.90        & 70.25       & 80.91      \\
Supervised*           & Food2K                                                                   & DeeplabV3                                                                       & Fine-tuning                    & 63.08    & 35.37    & 47.32    & 79.63       & 70.50        & 80.37      \\ 
BYOL*              & ImageNet-1K                                                              & DeeplabV3                                                                       & Fine-tuning                    & 56.25    & 28.39    & 40.20     & 75.65       & 66.02       & 76.63      \\
BYOL                    & Food2K                                                                   & DeeplabV3                                                                       & Fine-tuning                    & 60.86    & 31.72    & 43.91    & 82.08       & 72.56       & 83.28      \\
\rowcolor{green!20}BYOL+FeaSC                & Food2K                                                                   & DeeplabV3                                                                       & Fine-tuning                    & 64.14    & 36.22    & 48.87    & 82.61       & 74.12       & 84.14      \\ 
SimSiam*             & ImageNet-1K                                                              & DeeplabV3                                                                       & Fine-tuning                    & 59.54    & 30.93    & 42.66    & 73.46       & 60.15       & 72.65      \\
SimSiam                 & Food2K                                                                   & DeeplabV3                                                                       & Fine-tuning                    & 60.98    & 31.82    & 43.78    & 78.34       & 68.37       & 79.17      \\
\rowcolor{green!20}SimSiam+FeaSC             & Food2K                                                                   & DeeplabV3                                                                       & Fine-tuning                    & 63.86    & 35.68    & 48.20     & 82.62       & 73.98       & 83.70       \\ \hline
Supervised*           & ImageNet-1K                                                              & FCN                                                                             & Fine-tuning                    & 59.66    & 32.77    & 44.62    & 73.00          & 59.85       & 73.86      \\
Supervised*           & Food2K                                                                   & FCN                                                                             & Fine-tuning                    & 61.70     & 33.75    & 44.97    & 73.70        & 61.26       & 74.74      \\ 
BYOL*               & ImageNet-1K                                                              & FCN                                                                             & Fine-tuning                    & 57.59    & 27.93    & 39.31    & 67.49       & 52.39       & 67.78      \\
BYOL                    & Food2K                                                                   & FCN                                                                             & Fine-tuning                    & 59.08    & 30.80     & 42.36    & 73.28       & 59.54       & 73.45      \\
\rowcolor{green!20}BYOL+FeaSC                & Food2K                                                                   & FCN                                                                             & Fine-tuning                    & 61.30     & 32.36    & 44.46    & 76.54       & 65.14       & 78.03      \\ 
SimSiam*             & ImageNet-1K                                                              & FCN                                                                             & Fine-tuning                    & 59.54    & 30.93    & 42.66    & 71.29       & 56.40        & 70.91      \\
SimSiam                 & Food2K                                                                   & FCN                                                                             & Fine-tuning                    & 60.98    & 31.82    & 43.78    & 73.80        & 61.27       & 74.12      \\
\rowcolor{green!20}SimSiam+FeaSC             & Food2K                                                                   & FCN                                                                             & Fine-tuning                    & 62.32    & 34.49    & 46.20     & 76.74       & 65.07       & 77.80       \\ \bottomrule
\end{tabular}
\caption{Evaluation of food segmentation (\%) with different methods on FoodSeg 103 and UEC-FoodPix Complete.}
\label{tab:seg}
\vspace{-1.em}
\end{table*}

The proposed methods outperform their corresponding original self-supervised methods in both linear and fine-tuning evaluations. Specifically, when using the entire training data for linear evaluation, the proposed BYOL+FeaSC obtains performance gains over BYOL by 1.70\% on ETHZ Food-101, 3.27\% on Vireo Food-172, 4.98\% on ISIA Food-200 and 3.39\% on ISIA Food-500, while SimSiam+FeaSC improves SimSiam by 3.42\% on ETHZ Food-101, 4.69\% on Vireo Food-172, 4.97\% on ISIA Food-200 and 6.69\% on ISIA Food-500. 
When utilizing the complete training data for fine-tuning evaluation, BYOL+FeaSC outperforms BYOL by 0.86\% on Vireo Food-172, 0.55\% on ISIA Food-200, and 0.66\% on ISIA Food-500.
Noteworthy, the proposed BYOL+FeaSC shows a significant improvement over the supervised method in linear evaluation. 
When evaluated using the entire training data, the proposed BYOL+FeaSC outperforms the supervised method by 7.87\%, 4.69\%, 9.18\%, and 7.35\% on ETHZ Food-101 (73.49\% to 81.36\%), Vireo Food-172 (83.17\% to 87.86\%), ISIA Food-200 (54.86\% to 64.04\%), and ISIA Food-500 (48.59\% to 55.94\%), respectively. 
These experimental results demonstrate the high expressiveness of the features extracted by the proposed method.

The proposed method exhibits superior performance in linear evaluation with a small amount of training data.
As the amount of training data decreases, the advantages of the proposed method become increasingly apparent.
For example, on Vireo Food-172, using linear evaluation, SimSiam+FeaSC outperforms SimSiam by 4.69\% (76.89\% to 81.58\%) in the case of 100\% training data, 8.19\% (71.37\% to 79.56\%) in the 50\% case, 15.98\% (59.46\% to 75.44\%) in the 20\% case, and by 20.96\% (50.31\% to 71.27\%) in the 10\% case. These results demonstrate that the proposed method is capable of achieving high performance even with limited data. This characteristic is critical for practical use as procuring a significant quantity of labeled food images is arduous and costly.

\subsection{Food Segmentation}

 \textbf{Datasets.} Comparative experiments of food Segmentation are conducted on FoodSeg103 \cite{wu2021foodseg} and UEC-FoodPix Complete \citep{uecfoodpixcomplete}.
\textbf{FoodSeg103} is a western food segmentation dataset with 103 ingredient classes and 7,118 images, which includes 4,983 images for training and 2,135 image for testing.
\textbf{UEC-FoodPix Complete} is a released dataset of food image segmentation, which includes 9,000 images for training and 1,000 image for testing. The images are provided manually with pixel-wise 103 class labels. 

\textbf{Evaluation Protocols.} Similar to evaluation protocols of food recognition, the two standard evaluation protocols are used. 
The first involves training solely the segmentation head network while keeping the pre-trained weights frozen, known as head evaluation. The second protocol entails training the entire network, with the backbone initialized using pre-trained weights and is referred to as fine-tuning evaluation.

We adopt the segmentation methods FCN \citep{long2015fully} and DeepLabv3 \citep{chen2018encoder} for evaluation. We employ the same learning rate schedule (”poly” policy), momentum (0.9), and initial learning rate (0.01) as in previous works. The crop size was set to $512 \times 512$. To evaluate performance, we used mIOU (mean intersection over union), mAcc (mean accuracy of each class), and aAcc (accuracy for all pixels). mIoU is a standard measurement for semantic segmentation that evaluates the overlap and the union in inference and ground truth. mAcc is the mean accuracy of each class. aAcc is a more straightforward measurement that is the accuracy for all pixels.

\textbf{Experimental results.}
Table \ref{tab:seg} reports experimental results of different methods on FoodSeg103 and UEC-FoodPix Complete. 
The proposed methods exhibit superior performance when utilizing identical segmentation techniques in head evaluation compared to their original self-supervised methods. For example, equipped with DeeplabV3, the proposed BYOL+FeaSC improves BYOL by 3.59\% aAcc, 1.05\% mIoU, 1.72\% mACC on FoodSeg103, and its improvements are 4.77\% aAcc, 3.31\% mIoU, 5.93\% mACC on UEC-FoodPix Complete.
 The proposed SimSiam+FeaSC utilizing FCN improves SimSiam by 4.22\% aAcc, 1.16\% mIoU, 2.06\% mACC on FoodSeg103, and its improvements are 1.47\% aAcc, 0.92\% mIoU, 1.92\% mACC on UEC-FoodPix Complete.

In fine-tuning evaluation, the proposed methods outperform their original methods when identical segmentation techniques are utilized. 
For instance, equipped with DeeplabV3, the proposed SimSiam+FeaSC improves SimSiam by 2.88\% aAcc, 3.86\% mIoU, 4.42\% mACC on FoodSeg103, and its gains are 4.28\% aAcc, 5.61\% mIoU, 4.53\% mACC on UEC-FoodPix Complete.
Utilizing FCN, the proposed BYOL+FeaSC improves BYOL by 2.22\%  aAcc, 1.56\% mIoU, 2.10\% mACC on FoodSeg103, and its improvements are 3.26\% aAcc, 5.60\% mIoU, 4.58\% mACC on UEC-FoodPix Complete.
In addition, SimSiam+FeaSC combined with DeeplabV3 outperforms the supervised learning method by 0.88\% (47.32\% to 48.20\%) and 3.33\% (80.37\% to 83.70\%) mAcc on the FoodSeg103 and UEC-FoodPix Complete datasets, respectively. The corresponding results in combination with FCN are 1.23\% (44.97\% to 46.20\%) and 3.06\% (74.74\% to 77.80\%).
These experiments demonstrate the effectiveness of the proposed method on downstream segmentation tasks.

\subsection{Further Analysis}

\textbf{Ablation study.} To validate the effectiveness of the proposed method BYOL+FeaSC, we conducted ablation experiments and presented the results in Table \ref{tab:ablation}. The experiments demonstrate that feature suppression can significantly improve recognition accuracy. Furthermore, our proposed response-aware localization scheme can enhance the effect of feature suppression.
\begin{table}[h]
\begin{tabular}{ccccc}
\toprule
Method       & Food101        & Food172        & Food200        & Food500        \\\toprule
w/o S          & 79.66          & 84.59          & 59.06          & 52.55          \\
LRS          & 81.00          & 86.25          & 60.83          & 54.73          \\
RS           & 81.23          & 87.23          & 62.13          & 55.89          \\
\textbf{Our} & \textbf{81.36} & \textbf{87.86} & \textbf{64.04} & \textbf{55.94} \\ \bottomrule
\end{tabular}
\caption{Effect of different suppression strategies (\%). "w/o S": without feature suppression,  "LRS" : low-response suppression and "RS": random suppression.}
\label{tab:ablation}
\vspace{-1em}
\end{table}




\begin{figure}[h]
\setlength{\abovecaptionskip}{0cm}
  \centering
   \includegraphics[width=0.99\linewidth]{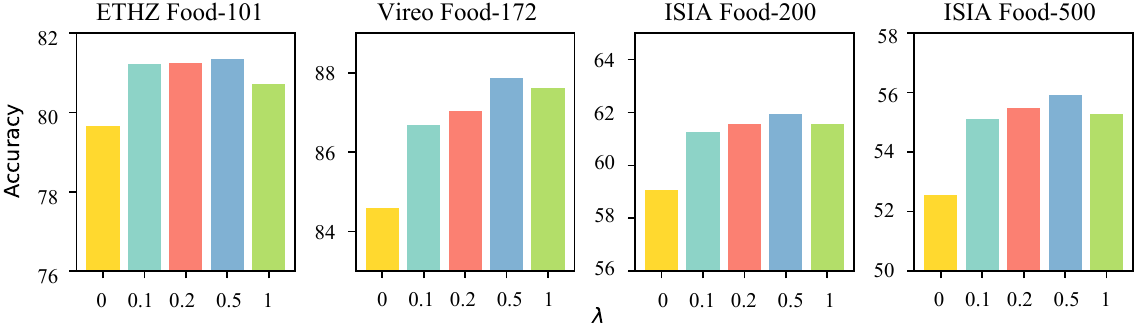}
    
   \caption{The recognition performance of the proposed method on four food datasets under different $\lambda$.}
   \label{fig:lambda}
   \vspace{-1em}
\end{figure}

\textbf{Analysis of the hyper-parameter $\lambda$.}
The hyper-parameter $\lambda$ in Eq. \ref{equ.final_loss} determines the balance between the original contrast term and the feature suppression contrast term. Figure \ref{fig:lambda} shows two key observations. First, the feature suppression contrast term can significantly improve recognition performance. Second, recognition performance is limited by a reverse U-shaped curve, with the optimal point at the top of the curve. This finding aligns with the theory of paper \cite{tian2020makes} that when two views’ mutual information is too high, it may introduce excessive noise that affects network generalization performance, conversely, when their mutual information is too low, there may not be enough signal to support network training.

\textbf{Computational complexity.}
We quantitatively compare the computational complexity of different suppression methods, and the results are shown in Table \ref{tab:comcom}.
It can be seen that neither the proposed FeaSC nor the direct image suppression adds almost no additional number of parameters. However, comparing MACs, it becomes clear that the proposed method adds almost no complexity, while going back to the image for suppression, MACs increase by 50\%. Self-supervised algorithms usually require a lot of computational resources, and from this perspective, the proposed method is clearly superior to the image level suppression.

 \begin{table}[]
\begin{tabular}{ccc}
\toprule
Method            & Params(M) & MACs(G) \\\hline
w/o Suppressing   & 74.2      & 16.58   \\
FeaSC             & 74.3      & 16.61   \\
Image Suppressing & 74.3      & 24.87  \\\toprule
\end{tabular}
\caption{Computational complexity of different methods.}
\label{tab:comcom}
\vspace{-2em}
\end{table}

\section{Conclusion and Future Work} \label{sec.con}
In this paper, we explore  self-supervised learning on food images and propose Feature-Suppressed Contrast (FeaSC) to boost self-supervised food pre-training by excluding comparisons of similar informative contents.
The proposed FeaSC leverages a response-aware scheme to identify  salient features in an unsupervised manner.
By suppressing some salient features in one view while leaving another contrast view unchanged, the mutual information between the two views decreases.
Consequently, the effectiveness of contrast learning for self-supervised food pre-training is improved.
Extensive qualitative and quantitative experiments have verified the effectiveness of the proposed method.
In future work, we will continue to investigate the performance of our method on other food datasets, such as Recipe1M. We will also further explore self-supervised pre-training methods, such as Dino, BarlowTwins, etc.

\bibliographystyle{ACM-Reference-Format}
\bibliography{ssl_food}


\begin{thebibliography}{48}


\ifx \showCODEN    \undefined \def \showCODEN     #1{\unskip}     \fi
\ifx \showDOI      \undefined \def \showDOI       #1{#1}\fi
\ifx \showISBNx    \undefined \def \showISBNx     #1{\unskip}     \fi
\ifx \showISBNxiii \undefined \def \showISBNxiii  #1{\unskip}     \fi
\ifx \showISSN     \undefined \def \showISSN      #1{\unskip}     \fi
\ifx \showLCCN     \undefined \def \showLCCN      #1{\unskip}     \fi
\ifx \shownote     \undefined \def \shownote      #1{#1}          \fi
\ifx \showarticletitle \undefined \def \showarticletitle #1{#1}   \fi
\ifx \showURL      \undefined \def \showURL       {\relax}        \fi
\providecommand\bibfield[2]{#2}
\providecommand\bibinfo[2]{#2}
\providecommand\natexlab[1]{#1}
\providecommand\showeprint[2][]{arXiv:#2}

\bibitem[\protect\citeauthoryear{Assran, Caron, Misra, Bojanowski, Bordes,
  Vincent, Joulin, Rabbat, and Ballas}{Assran et~al\mbox{.}}{2022}]%
        {assran2022masked}
\bibfield{author}{\bibinfo{person}{Mahmoud Assran}, \bibinfo{person}{Mathilde
  Caron}, \bibinfo{person}{Ishan Misra}, \bibinfo{person}{Piotr Bojanowski},
  \bibinfo{person}{Florian Bordes}, \bibinfo{person}{Pascal Vincent},
  \bibinfo{person}{Armand Joulin}, \bibinfo{person}{Mike Rabbat}, {and}
  \bibinfo{person}{Nicolas Ballas}.} \bibinfo{year}{2022}\natexlab{}.
\newblock \showarticletitle{Masked siamese networks for label-efficient
  learning}. In \bibinfo{booktitle}{\emph{ECCV}}. \bibinfo{pages}{456--473}.
\newblock


\bibitem[\protect\citeauthoryear{Bao, Dong, Piao, and Wei}{Bao
  et~al\mbox{.}}{2021}]%
        {bao2021beit}
\bibfield{author}{\bibinfo{person}{Hangbo Bao}, \bibinfo{person}{Li Dong},
  \bibinfo{person}{Songhao Piao}, {and} \bibinfo{person}{Furu Wei}.}
  \bibinfo{year}{2021}\natexlab{}.
\newblock \showarticletitle{BEiT: BERT Pre-Training of Image Transformers}. In
  \bibinfo{booktitle}{\emph{ICLR}}.
\newblock


\bibitem[\protect\citeauthoryear{Bardes, Ponce, and LeCun}{Bardes
  et~al\mbox{.}}{2022}]%
        {bardes2022variance}
\bibfield{author}{\bibinfo{person}{Adrien Bardes}, \bibinfo{person}{Jean
  Ponce}, {and} \bibinfo{person}{Yann LeCun}.} \bibinfo{year}{2022}\natexlab{}.
\newblock \showarticletitle{Variance-invariance-covariance regularization for
  self-supervised learning}. In \bibinfo{booktitle}{\emph{ICLR}}.
\newblock


\bibitem[\protect\citeauthoryear{Bossard, Guillaumin, and Van~Gool}{Bossard
  et~al\mbox{.}}{2014}]%
        {Bossard-Food101-ECCV2014}
\bibfield{author}{\bibinfo{person}{Lukas Bossard}, \bibinfo{person}{Matthieu
  Guillaumin}, {and} \bibinfo{person}{Luc Van~Gool}.}
  \bibinfo{year}{2014}\natexlab{}.
\newblock \showarticletitle{Food-101--mining discriminative components with
  random forests}. In \bibinfo{booktitle}{\emph{ECCV}}.
  \bibinfo{pages}{446--461}.
\newblock


\bibitem[\protect\citeauthoryear{Brosnan and Sun}{Brosnan and Sun}{2004}]%
        {brosnan2004improving}
\bibfield{author}{\bibinfo{person}{Tadhg Brosnan} {and} \bibinfo{person}{Da-Wen
  Sun}.} \bibinfo{year}{2004}\natexlab{}.
\newblock \showarticletitle{Improving quality inspection of food products by
  computer vision----a review}.
\newblock \bibinfo{journal}{\emph{Journal of food engineering}}
  \bibinfo{volume}{61}, \bibinfo{number}{1} (\bibinfo{year}{2004}),
  \bibinfo{pages}{3--16}.
\newblock


\bibitem[\protect\citeauthoryear{Caron, Misra, Mairal, Goyal, Bojanowski, and
  Joulin}{Caron et~al\mbox{.}}{2020}]%
        {caron2020unsupervised}
\bibfield{author}{\bibinfo{person}{Mathilde Caron}, \bibinfo{person}{Ishan
  Misra}, \bibinfo{person}{Julien Mairal}, \bibinfo{person}{Priya Goyal},
  \bibinfo{person}{Piotr Bojanowski}, {and} \bibinfo{person}{Armand Joulin}.}
  \bibinfo{year}{2020}\natexlab{}.
\newblock \showarticletitle{Unsupervised learning of visual features by
  contrasting cluster assignments}. In \bibinfo{booktitle}{\emph{NeurIPS}},
  Vol.~\bibinfo{volume}{33}. \bibinfo{pages}{9912--9924}.
\newblock


\bibitem[\protect\citeauthoryear{Caron, Touvron, Misra, J{\'e}gou, Mairal,
  Bojanowski, and Joulin}{Caron et~al\mbox{.}}{2021}]%
        {caron2021emerging}
\bibfield{author}{\bibinfo{person}{Mathilde Caron}, \bibinfo{person}{Hugo
  Touvron}, \bibinfo{person}{Ishan Misra}, \bibinfo{person}{Herv{\'e}
  J{\'e}gou}, \bibinfo{person}{Julien Mairal}, \bibinfo{person}{Piotr
  Bojanowski}, {and} \bibinfo{person}{Armand Joulin}.}
  \bibinfo{year}{2021}\natexlab{}.
\newblock \showarticletitle{Emerging properties in self-supervised vision
  transformers}. In \bibinfo{booktitle}{\emph{ICCV}}.
  \bibinfo{pages}{9650--9660}.
\newblock


\bibitem[\protect\citeauthoryear{Chen and Ngo}{Chen and Ngo}{2016}]%
        {Chen-DIRCRR-MM2016}
\bibfield{author}{\bibinfo{person}{Jingjing Chen} {and}
  \bibinfo{person}{Chong-Wah Ngo}.} \bibinfo{year}{2016}\natexlab{}.
\newblock \showarticletitle{Deep-based ingredient recognition for cooking
  recipe retrieval}. In \bibinfo{booktitle}{\emph{ACM MM}}.
  \bibinfo{pages}{32--41}.
\newblock


\bibitem[\protect\citeauthoryear{Chen, Zhu, Papandreou, Schroff, and Adam}{Chen
  et~al\mbox{.}}{2018}]%
        {chen2018encoder}
\bibfield{author}{\bibinfo{person}{Liang-Chieh Chen}, \bibinfo{person}{Yukun
  Zhu}, \bibinfo{person}{George Papandreou}, \bibinfo{person}{Florian Schroff},
  {and} \bibinfo{person}{Hartwig Adam}.} \bibinfo{year}{2018}\natexlab{}.
\newblock \showarticletitle{Encoder-decoder with atrous separable convolution
  for semantic image segmentation}. In \bibinfo{booktitle}{\emph{ECCV}}.
  \bibinfo{pages}{801--818}.
\newblock


\bibitem[\protect\citeauthoryear{Chen, Kornblith, Norouzi, and Hinton}{Chen
  et~al\mbox{.}}{2020}]%
        {chen2020simple}
\bibfield{author}{\bibinfo{person}{Ting Chen}, \bibinfo{person}{Simon
  Kornblith}, \bibinfo{person}{Mohammad Norouzi}, {and}
  \bibinfo{person}{Geoffrey Hinton}.} \bibinfo{year}{2020}\natexlab{}.
\newblock \showarticletitle{A simple framework for contrastive learning of
  visual representations}. In \bibinfo{booktitle}{\emph{ICML}}.
  \bibinfo{pages}{1597--1607}.
\newblock


\bibitem[\protect\citeauthoryear{Chen and Yu}{Chen and Yu}{2021}]%
        {chen2021review}
\bibfield{author}{\bibinfo{person}{Tzu-Chia Chen} {and}
  \bibinfo{person}{Shu-Yan Yu}.} \bibinfo{year}{2021}\natexlab{}.
\newblock \showarticletitle{The review of food safety inspection system based
  on artificial intelligence, image processing, and robotic}.
\newblock \bibinfo{journal}{\emph{Food Science and Technology}}
  \bibinfo{volume}{42} (\bibinfo{year}{2021}).
\newblock


\bibitem[\protect\citeauthoryear{Chen and He}{Chen and He}{2021}]%
        {chen2021exploring}
\bibfield{author}{\bibinfo{person}{Xinlei Chen} {and} \bibinfo{person}{Kaiming
  He}.} \bibinfo{year}{2021}\natexlab{}.
\newblock \showarticletitle{Exploring simple siamese representation learning}.
  In \bibinfo{booktitle}{\emph{CVPR}}. \bibinfo{pages}{15750--15758}.
\newblock


\bibitem[\protect\citeauthoryear{Cheng, Lai, Gao, and Han}{Cheng
  et~al\mbox{.}}{2023}]%
        {cheng2023class}
\bibfield{author}{\bibinfo{person}{Gong Cheng}, \bibinfo{person}{Pujian Lai},
  \bibinfo{person}{Decheng Gao}, {and} \bibinfo{person}{Junwei Han}.}
  \bibinfo{year}{2023}\natexlab{}.
\newblock \showarticletitle{Class attention network for image recognition}.
\newblock \bibinfo{journal}{\emph{Science China Information Sciences}}
  \bibinfo{volume}{66}, \bibinfo{number}{3} (\bibinfo{year}{2023}),
  \bibinfo{pages}{132105}.
\newblock


\bibitem[\protect\citeauthoryear{Choe and Shim}{Choe and Shim}{2019}]%
        {choe2019attention}
\bibfield{author}{\bibinfo{person}{Junsuk Choe} {and} \bibinfo{person}{Hyunjung
  Shim}.} \bibinfo{year}{2019}\natexlab{}.
\newblock \showarticletitle{Attention-based dropout layer for weakly supervised
  object localization}. In \bibinfo{booktitle}{\emph{Proceedings of the
  IEEE/CVF Conference on Computer Vision and Pattern Recognition}}.
  \bibinfo{pages}{2219--2228}.
\newblock


\bibitem[\protect\citeauthoryear{Fan, Liu, Chen, Zhang, and Gan}{Fan
  et~al\mbox{.}}{2021}]%
        {fan2021does}
\bibfield{author}{\bibinfo{person}{Lijie Fan}, \bibinfo{person}{Sijia Liu},
  \bibinfo{person}{Pin-Yu Chen}, \bibinfo{person}{Gaoyuan Zhang}, {and}
  \bibinfo{person}{Chuang Gan}.} \bibinfo{year}{2021}\natexlab{}.
\newblock \showarticletitle{When does contrastive learning preserve adversarial
  robustness from pretraining to finetuning?}. In
  \bibinfo{booktitle}{\emph{NeurIPS}}, Vol.~\bibinfo{volume}{34}.
  \bibinfo{pages}{21480--21492}.
\newblock


\bibitem[\protect\citeauthoryear{Grill, Strub, Altch{\'e}, Tallec, Richemond,
  Buchatskaya, Doersch, Avila~Pires, Guo, Gheshlaghi~Azar, et~al\mbox{.}}{Grill
  et~al\mbox{.}}{2020}]%
        {grill2020bootstrap}
\bibfield{author}{\bibinfo{person}{Jean-Bastien Grill},
  \bibinfo{person}{Florian Strub}, \bibinfo{person}{Florent Altch{\'e}},
  \bibinfo{person}{Corentin Tallec}, \bibinfo{person}{Pierre Richemond},
  \bibinfo{person}{Elena Buchatskaya}, \bibinfo{person}{Carl Doersch},
  \bibinfo{person}{Bernardo Avila~Pires}, \bibinfo{person}{Zhaohan Guo},
  \bibinfo{person}{Mohammad Gheshlaghi~Azar}, {et~al\mbox{.}}}
  \bibinfo{year}{2020}\natexlab{}.
\newblock \showarticletitle{Bootstrap your own latent-a new approach to
  self-supervised learning}. In \bibinfo{booktitle}{\emph{NeurIPS}},
  Vol.~\bibinfo{volume}{33}. \bibinfo{pages}{21271--21284}.
\newblock


\bibitem[\protect\citeauthoryear{He, Chen, Xie, Li, Doll{\'a}r, and
  Girshick}{He et~al\mbox{.}}{2022}]%
        {he2022masked}
\bibfield{author}{\bibinfo{person}{Kaiming He}, \bibinfo{person}{Xinlei Chen},
  \bibinfo{person}{Saining Xie}, \bibinfo{person}{Yanghao Li},
  \bibinfo{person}{Piotr Doll{\'a}r}, {and} \bibinfo{person}{Ross Girshick}.}
  \bibinfo{year}{2022}\natexlab{}.
\newblock \showarticletitle{Masked autoencoders are scalable vision learners}.
  In \bibinfo{booktitle}{\emph{CVPR}}. \bibinfo{pages}{16000--16009}.
\newblock


\bibitem[\protect\citeauthoryear{He, Fan, Wu, Xie, and Girshick}{He
  et~al\mbox{.}}{2020}]%
        {he2020momentum}
\bibfield{author}{\bibinfo{person}{Kaiming He}, \bibinfo{person}{Haoqi Fan},
  \bibinfo{person}{Yuxin Wu}, \bibinfo{person}{Saining Xie}, {and}
  \bibinfo{person}{Ross Girshick}.} \bibinfo{year}{2020}\natexlab{}.
\newblock \showarticletitle{Momentum contrast for unsupervised visual
  representation learning}. In \bibinfo{booktitle}{\emph{CVPR}}.
  \bibinfo{pages}{9729--9738}.
\newblock


\bibitem[\protect\citeauthoryear{Hu, Wang, Hu, and Qi}{Hu
  et~al\mbox{.}}{2021}]%
        {hu2021adco}
\bibfield{author}{\bibinfo{person}{Qianjiang Hu}, \bibinfo{person}{Xiao Wang},
  \bibinfo{person}{Wei Hu}, {and} \bibinfo{person}{Guo-Jun Qi}.}
  \bibinfo{year}{2021}\natexlab{}.
\newblock \showarticletitle{Adco: Adversarial contrast for efficient learning
  of unsupervised representations from self-trained negative adversaries}. In
  \bibinfo{booktitle}{\emph{CVPR}}. \bibinfo{pages}{1074--1083}.
\newblock


\bibitem[\protect\citeauthoryear{Hua, Wang, Xue, Ren, Wang, and Zhao}{Hua
  et~al\mbox{.}}{2021}]%
        {hua2021feature}
\bibfield{author}{\bibinfo{person}{Tianyu Hua}, \bibinfo{person}{Wenxiao Wang},
  \bibinfo{person}{Zihui Xue}, \bibinfo{person}{Sucheng Ren},
  \bibinfo{person}{Yue Wang}, {and} \bibinfo{person}{Hang Zhao}.}
  \bibinfo{year}{2021}\natexlab{}.
\newblock \showarticletitle{On feature decorrelation in self-supervised
  learning}. In \bibinfo{booktitle}{\emph{CVPR}}. \bibinfo{pages}{9598--9608}.
\newblock


\bibitem[\protect\citeauthoryear{Jiang, Qiu, Liu, Huang, and Lin}{Jiang
  et~al\mbox{.}}{2020}]%
        {jiang2020deepfood}
\bibfield{author}{\bibinfo{person}{Landu Jiang}, \bibinfo{person}{Bojia Qiu},
  \bibinfo{person}{Xue Liu}, \bibinfo{person}{Chenxi Huang}, {and}
  \bibinfo{person}{Kunhui Lin}.} \bibinfo{year}{2020}\natexlab{}.
\newblock \showarticletitle{DeepFood: food image analysis and dietary
  assessment via deep model}.
\newblock \bibinfo{journal}{\emph{IEEE Access}}  \bibinfo{volume}{8}
  (\bibinfo{year}{2020}), \bibinfo{pages}{47477--47489}.
\newblock


\bibitem[\protect\citeauthoryear{Kim, Tack, and Hwang}{Kim
  et~al\mbox{.}}{2020}]%
        {kim2020adversarial}
\bibfield{author}{\bibinfo{person}{Minseon Kim}, \bibinfo{person}{Jihoon Tack},
  {and} \bibinfo{person}{Sung~Ju Hwang}.} \bibinfo{year}{2020}\natexlab{}.
\newblock \showarticletitle{Adversarial self-supervised contrastive learning}.
  In \bibinfo{booktitle}{\emph{NeurIPS}}, Vol.~\bibinfo{volume}{33}.
  \bibinfo{pages}{2983--2994}.
\newblock


\bibitem[\protect\citeauthoryear{Li, Zhou, Xiong, and Hoi}{Li
  et~al\mbox{.}}{[n.d.]}]%
        {liprototypical}
\bibfield{author}{\bibinfo{person}{Junnan Li}, \bibinfo{person}{Pan Zhou},
  \bibinfo{person}{Caiming Xiong}, {and} \bibinfo{person}{Steven Hoi}.}
  \bibinfo{year}{[n.d.]}\natexlab{}.
\newblock \showarticletitle{Prototypical Contrastive Learning of Unsupervised
  Representations}. In \bibinfo{booktitle}{\emph{ICLR}}.
\newblock


\bibitem[\protect\citeauthoryear{Li, Chang, Ma, Shan, and Xilin}{Li
  et~al\mbox{.}}{2022}]%
        {lioptimal}
\bibfield{author}{\bibinfo{person}{Yinqi Li}, \bibinfo{person}{Hong Chang},
  \bibinfo{person}{Bingpeng Ma}, \bibinfo{person}{Shiguang Shan}, {and}
  \bibinfo{person}{CHEN Xilin}.} \bibinfo{year}{2022}\natexlab{}.
\newblock \showarticletitle{Optimal Positive Generation via Latent
  Transformation for Contrastive Learning}. In
  \bibinfo{booktitle}{\emph{NeurIPS}}.
\newblock


\bibitem[\protect\citeauthoryear{Long, Shelhamer, and Darrell}{Long
  et~al\mbox{.}}{2015}]%
        {long2015fully}
\bibfield{author}{\bibinfo{person}{Jonathan Long}, \bibinfo{person}{Evan
  Shelhamer}, {and} \bibinfo{person}{Trevor Darrell}.}
  \bibinfo{year}{2015}\natexlab{}.
\newblock \showarticletitle{Fully convolutional networks for semantic
  segmentation}. In \bibinfo{booktitle}{\emph{CVPR}}.
  \bibinfo{pages}{3431--3440}.
\newblock


\bibitem[\protect\citeauthoryear{Min, Jiang, and Jain}{Min
  et~al\mbox{.}}{2019a}]%
        {min2019food}
\bibfield{author}{\bibinfo{person}{Weiqing Min}, \bibinfo{person}{Shuqiang
  Jiang}, {and} \bibinfo{person}{Ramesh Jain}.}
  \bibinfo{year}{2019}\natexlab{a}.
\newblock \showarticletitle{Food recommendation: Framework, existing solutions,
  and challenges}.
\newblock \bibinfo{journal}{\emph{IEEE TMM}} \bibinfo{volume}{22},
  \bibinfo{number}{10} (\bibinfo{year}{2019}), \bibinfo{pages}{2659--2671}.
\newblock


\bibitem[\protect\citeauthoryear{Min, Liu, Luo, and Jiang}{Min
  et~al\mbox{.}}{2019b}]%
        {min2019ingredient}
\bibfield{author}{\bibinfo{person}{Weiqing Min}, \bibinfo{person}{Linhu Liu},
  \bibinfo{person}{Zhengdong Luo}, {and} \bibinfo{person}{Shuqiang Jiang}.}
  \bibinfo{year}{2019}\natexlab{b}.
\newblock \showarticletitle{Ingredient-Guided Cascaded Multi-Attention Network
  for Food Recognition}. In \bibinfo{booktitle}{\emph{ACM MM}}.
  \bibinfo{publisher}{{ACM}}, \bibinfo{pages}{1331--1339}.
\newblock


\bibitem[\protect\citeauthoryear{Min, Liu, Wang, Luo, Wei, Wei, and Jiang}{Min
  et~al\mbox{.}}{2020}]%
        {min2020isia}
\bibfield{author}{\bibinfo{person}{Weiqing Min}, \bibinfo{person}{Linhu Liu},
  \bibinfo{person}{Zhiling Wang}, \bibinfo{person}{Zhengdong Luo},
  \bibinfo{person}{Xiaoming Wei}, \bibinfo{person}{Xiaolin Wei}, {and}
  \bibinfo{person}{Shuqiang Jiang}.} \bibinfo{year}{2020}\natexlab{}.
\newblock \showarticletitle{Isia food-500: A dataset for large-scale food
  recognition via stacked global-local attention network}. In
  \bibinfo{booktitle}{\emph{ACM MM}}. \bibinfo{pages}{393--401}.
\newblock


\bibitem[\protect\citeauthoryear{Min, Wang, Liu, Luo, Kang, Wei, Wei, and
  Jiang}{Min et~al\mbox{.}}{2023}]%
        {min2023large}
\bibfield{author}{\bibinfo{person}{Weiqing Min}, \bibinfo{person}{Zhiling
  Wang}, \bibinfo{person}{Yuxin Liu}, \bibinfo{person}{Mengjiang Luo},
  \bibinfo{person}{Liping Kang}, \bibinfo{person}{Xiaoming Wei},
  \bibinfo{person}{Xiaolin Wei}, {and} \bibinfo{person}{Shuqiang Jiang}.}
  \bibinfo{year}{2023}\natexlab{}.
\newblock \showarticletitle{Large scale visual food recognition}.
\newblock \bibinfo{journal}{\emph{IEEE TPAMI}} (\bibinfo{year}{2023}).
\newblock


\bibitem[\protect\citeauthoryear{Okamoto and Yanai}{Okamoto and Yanai}{2021}]%
        {uecfoodpixcomplete}
\bibfield{author}{\bibinfo{person}{Kaimu Okamoto} {and} \bibinfo{person}{Keiji
  Yanai}.} \bibinfo{year}{2021}\natexlab{}.
\newblock \showarticletitle{{UEC-FoodPIX Complete}: A Large-scale Food Image
  Segmentation Dataset}. In \bibinfo{booktitle}{\emph{ICPRW}}.
\newblock


\bibitem[\protect\citeauthoryear{Oord, Li, and Vinyals}{Oord
  et~al\mbox{.}}{2018}]%
        {oord2018representation}
\bibfield{author}{\bibinfo{person}{Aaron van~den Oord}, \bibinfo{person}{Yazhe
  Li}, {and} \bibinfo{person}{Oriol Vinyals}.} \bibinfo{year}{2018}\natexlab{}.
\newblock \showarticletitle{Representation learning with contrastive predictive
  coding}.
\newblock \bibinfo{journal}{\emph{arXiv preprint arXiv:1807.03748}}
  (\bibinfo{year}{2018}).
\newblock


\bibitem[\protect\citeauthoryear{Peng, Wang, Zhu, Wang, and You}{Peng
  et~al\mbox{.}}{2022}]%
        {peng2022crafting}
\bibfield{author}{\bibinfo{person}{Xiangyu Peng}, \bibinfo{person}{Kai Wang},
  \bibinfo{person}{Zheng Zhu}, \bibinfo{person}{Mang Wang}, {and}
  \bibinfo{person}{Yang You}.} \bibinfo{year}{2022}\natexlab{}.
\newblock \showarticletitle{Crafting better contrastive views for siamese
  representation learning}. In \bibinfo{booktitle}{\emph{CVPR}}.
  \bibinfo{pages}{16031--16040}.
\newblock


\bibitem[\protect\citeauthoryear{Selvaraju, Cogswell, Das, Vedantam, Parikh,
  and Batra}{Selvaraju et~al\mbox{.}}{2017}]%
        {selvaraju2017grad}
\bibfield{author}{\bibinfo{person}{Ramprasaath~R Selvaraju},
  \bibinfo{person}{Michael Cogswell}, \bibinfo{person}{Abhishek Das},
  \bibinfo{person}{Ramakrishna Vedantam}, \bibinfo{person}{Devi Parikh}, {and}
  \bibinfo{person}{Dhruv Batra}.} \bibinfo{year}{2017}\natexlab{}.
\newblock \showarticletitle{Grad-cam: Visual explanations from deep networks
  via gradient-based localization}. In \bibinfo{booktitle}{\emph{Proceedings of
  the IEEE international conference on computer vision}}.
  \bibinfo{pages}{618--626}.
\newblock


\bibitem[\protect\citeauthoryear{Shen, Sun, Wei, Jiang, and Yang}{Shen
  et~al\mbox{.}}{2022}]%
        {shen2022semicon}
\bibfield{author}{\bibinfo{person}{Yang Shen}, \bibinfo{person}{Xuhao Sun},
  \bibinfo{person}{Xiu-Shen Wei}, \bibinfo{person}{Qing-Yuan Jiang}, {and}
  \bibinfo{person}{Jian Yang}.} \bibinfo{year}{2022}\natexlab{}.
\newblock \showarticletitle{SEMICON: A Learning-to-Hash Solution for
  Large-Scale Fine-Grained Image Retrieval}. In
  \bibinfo{booktitle}{\emph{European Conference on Computer Vision}}. Springer,
  \bibinfo{pages}{531--548}.
\newblock


\bibitem[\protect\citeauthoryear{Shi, Siddharth, Torr, and Kosiorek}{Shi
  et~al\mbox{.}}{2022}]%
        {shi2022adversarial}
\bibfield{author}{\bibinfo{person}{Yuge Shi}, \bibinfo{person}{N Siddharth},
  \bibinfo{person}{Philip Torr}, {and} \bibinfo{person}{Adam~R Kosiorek}.}
  \bibinfo{year}{2022}\natexlab{}.
\newblock \showarticletitle{Adversarial masking for self-supervised learning}.
  In \bibinfo{booktitle}{\emph{ICML}}. \bibinfo{pages}{20026--20040}.
\newblock


\bibitem[\protect\citeauthoryear{Sun and Li}{Sun and Li}{2023}]%
        {sun2023enhancing}
\bibfield{author}{\bibinfo{person}{Hui Sun} {and} \bibinfo{person}{Ming Li}.}
  \bibinfo{year}{2023}\natexlab{}.
\newblock \showarticletitle{Enhancing unsupervised domain adaptation by
  exploiting the conceptual consistency of multiple self-supervised tasks}.
\newblock \bibinfo{journal}{\emph{Science China Information Sciences}}
  \bibinfo{volume}{66}, \bibinfo{number}{4} (\bibinfo{year}{2023}),
  \bibinfo{pages}{142101}.
\newblock


\bibitem[\protect\citeauthoryear{Tian, Sun, Poole, Krishnan, Schmid, and
  Isola}{Tian et~al\mbox{.}}{2020}]%
        {tian2020makes}
\bibfield{author}{\bibinfo{person}{Yonglong Tian}, \bibinfo{person}{Chen Sun},
  \bibinfo{person}{Ben Poole}, \bibinfo{person}{Dilip Krishnan},
  \bibinfo{person}{Cordelia Schmid}, {and} \bibinfo{person}{Phillip Isola}.}
  \bibinfo{year}{2020}\natexlab{}.
\newblock \showarticletitle{What makes for good views for contrastive
  learning?}. In \bibinfo{booktitle}{\emph{NeurIPS}},
  Vol.~\bibinfo{volume}{33}. \bibinfo{pages}{6827--6839}.
\newblock


\bibitem[\protect\citeauthoryear{Wang, Min, Li, Dong, Li, and Jiang}{Wang
  et~al\mbox{.}}{2022a}]%
        {wang2022review}
\bibfield{author}{\bibinfo{person}{Wei Wang}, \bibinfo{person}{Weiqing Min},
  \bibinfo{person}{Tianhao Li}, \bibinfo{person}{Xiaoxiao Dong},
  \bibinfo{person}{Haisheng Li}, {and} \bibinfo{person}{Shuqiang Jiang}.}
  \bibinfo{year}{2022}\natexlab{a}.
\newblock \showarticletitle{A review on vision-based analysis for automatic
  dietary assessment}.
\newblock \bibinfo{journal}{\emph{Trends in Food Science \& Technology}}
  (\bibinfo{year}{2022}).
\newblock


\bibitem[\protect\citeauthoryear{Wang, Min, Li, Kang, Wei, Wei, and Jiang}{Wang
  et~al\mbox{.}}{2022b}]%
        {wang2022ingredient}
\bibfield{author}{\bibinfo{person}{Zhiling Wang}, \bibinfo{person}{Weiqing
  Min}, \bibinfo{person}{Zhuo Li}, \bibinfo{person}{Liping Kang},
  \bibinfo{person}{Xiaoming Wei}, \bibinfo{person}{Xiaolin Wei}, {and}
  \bibinfo{person}{Shuqiang Jiang}.} \bibinfo{year}{2022}\natexlab{b}.
\newblock \showarticletitle{Ingredient-Guided Region Discovery and Relationship
  Modeling for Food Category-Ingredient Prediction}.
\newblock \bibinfo{journal}{\emph{IEEE TIP}}  \bibinfo{volume}{31}
  (\bibinfo{year}{2022}), \bibinfo{pages}{5214--5226}.
\newblock


\bibitem[\protect\citeauthoryear{Wei, Fan, Xie, Wu, Yuille, and
  Feichtenhofer}{Wei et~al\mbox{.}}{2022}]%
        {wei2022masked}
\bibfield{author}{\bibinfo{person}{Chen Wei}, \bibinfo{person}{Haoqi Fan},
  \bibinfo{person}{Saining Xie}, \bibinfo{person}{Chao-Yuan Wu},
  \bibinfo{person}{Alan Yuille}, {and} \bibinfo{person}{Christoph
  Feichtenhofer}.} \bibinfo{year}{2022}\natexlab{}.
\newblock \showarticletitle{Masked feature prediction for self-supervised
  visual pre-training}. In \bibinfo{booktitle}{\emph{CVPR}}.
  \bibinfo{pages}{14668--14678}.
\newblock


\bibitem[\protect\citeauthoryear{Wei, Luo, Wu, and Zhou}{Wei
  et~al\mbox{.}}{2017}]%
        {wei2017selective}
\bibfield{author}{\bibinfo{person}{Xiu-Shen Wei}, \bibinfo{person}{Jian-Hao
  Luo}, \bibinfo{person}{Jianxin Wu}, {and} \bibinfo{person}{Zhi-Hua Zhou}.}
  \bibinfo{year}{2017}\natexlab{}.
\newblock \showarticletitle{Selective convolutional descriptor aggregation for
  fine-grained image retrieval}.
\newblock \bibinfo{journal}{\emph{IEEE transactions on image processing}}
  \bibinfo{volume}{26}, \bibinfo{number}{6} (\bibinfo{year}{2017}),
  \bibinfo{pages}{2868--2881}.
\newblock


\bibitem[\protect\citeauthoryear{Wu, Fu, Liu, Lim, Hoi, and Sun}{Wu
  et~al\mbox{.}}{2021}]%
        {wu2021foodseg}
\bibfield{author}{\bibinfo{person}{Xiongwei Wu}, \bibinfo{person}{Xin Fu},
  \bibinfo{person}{Ying Liu}, \bibinfo{person}{Ee-Peng Lim},
  \bibinfo{person}{Steven~CH Hoi}, {and} \bibinfo{person}{Qianru Sun}.}
  \bibinfo{year}{2021}\natexlab{}.
\newblock \showarticletitle{A Large-Scale Benchmark for Food Image
  Segmentation}. In \bibinfo{booktitle}{\emph{ACM MM}}.
\newblock


\bibitem[\protect\citeauthoryear{Xie, Zhang, Cao, Lin, Bao, Yao, Dai, and
  Hu}{Xie et~al\mbox{.}}{2022}]%
        {xie2022simmim}
\bibfield{author}{\bibinfo{person}{Zhenda Xie}, \bibinfo{person}{Zheng Zhang},
  \bibinfo{person}{Yue Cao}, \bibinfo{person}{Yutong Lin},
  \bibinfo{person}{Jianmin Bao}, \bibinfo{person}{Zhuliang Yao},
  \bibinfo{person}{Qi Dai}, {and} \bibinfo{person}{Han Hu}.}
  \bibinfo{year}{2022}\natexlab{}.
\newblock \showarticletitle{Simmim: A simple framework for masked image
  modeling}. In \bibinfo{booktitle}{\emph{CVPR}}. \bibinfo{pages}{9653--9663}.
\newblock


\bibitem[\protect\citeauthoryear{Yan, Tan, and Tai}{Yan et~al\mbox{.}}{2022}]%
        {yan2022repeatable}
\bibfield{author}{\bibinfo{person}{Pei Yan}, \bibinfo{person}{Yihua Tan}, {and}
  \bibinfo{person}{Yuan Tai}.} \bibinfo{year}{2022}\natexlab{}.
\newblock \showarticletitle{Repeatable adaptive keypoint detection via
  self-supervised learning}.
\newblock \bibinfo{journal}{\emph{Science China Information Sciences}}
  \bibinfo{volume}{65}, \bibinfo{number}{11} (\bibinfo{year}{2022}),
  \bibinfo{pages}{212103}.
\newblock


\bibitem[\protect\citeauthoryear{Zbontar, Jing, Misra, LeCun, and Deny}{Zbontar
  et~al\mbox{.}}{2021}]%
        {zbontar2021barlow}
\bibfield{author}{\bibinfo{person}{Jure Zbontar}, \bibinfo{person}{Li Jing},
  \bibinfo{person}{Ishan Misra}, \bibinfo{person}{Yann LeCun}, {and}
  \bibinfo{person}{St{\'e}phane Deny}.} \bibinfo{year}{2021}\natexlab{}.
\newblock \showarticletitle{Barlow twins: Self-supervised learning via
  redundancy reduction}. In \bibinfo{booktitle}{\emph{ICML}}.
  \bibinfo{pages}{12310--12320}.
\newblock


\bibitem[\protect\citeauthoryear{Zhou, Khosla, Lapedriza, Oliva, and
  Torralba}{Zhou et~al\mbox{.}}{2016}]%
        {zhou2016learning}
\bibfield{author}{\bibinfo{person}{Bolei Zhou}, \bibinfo{person}{Aditya
  Khosla}, \bibinfo{person}{Agata Lapedriza}, \bibinfo{person}{Aude Oliva},
  {and} \bibinfo{person}{Antonio Torralba}.} \bibinfo{year}{2016}\natexlab{}.
\newblock \showarticletitle{Learning deep features for discriminative
  localization}. In \bibinfo{booktitle}{\emph{Proceedings of the IEEE
  conference on computer vision and pattern recognition}}.
  \bibinfo{pages}{2921--2929}.
\newblock


\bibitem[\protect\citeauthoryear{Zhou, Wei, Wang, Shen, Xie, Yuille, and
  Kong}{Zhou et~al\mbox{.}}{2021}]%
        {zhou2022ibot}
\bibfield{author}{\bibinfo{person}{Jinghao Zhou}, \bibinfo{person}{Chen Wei},
  \bibinfo{person}{Huiyu Wang}, \bibinfo{person}{Wei Shen},
  \bibinfo{person}{Cihang Xie}, \bibinfo{person}{Alan Yuille}, {and}
  \bibinfo{person}{Tao Kong}.} \bibinfo{year}{2021}\natexlab{}.
\newblock \showarticletitle{ibot: Image bert pre-training with online
  tokenizer}. In \bibinfo{booktitle}{\emph{ICLR}}.
\newblock


\bibitem[\protect\citeauthoryear{Zhu, Liu, and Tian}{Zhu et~al\mbox{.}}{2023}]%
        {zhu2023learn}
\bibfield{author}{\bibinfo{person}{Yaohui Zhu}, \bibinfo{person}{Linhu Liu},
  {and} \bibinfo{person}{Jiang Tian}.} \bibinfo{year}{2023}\natexlab{}.
\newblock \showarticletitle{Learn More for Food Recognition via Progressive
  Self-Distillation}. In \bibinfo{booktitle}{\emph{AAAI}}.
\newblock


\end{thebibliography}

\end{document}